\newcommand{\ourmodel}{CoordAR}

\documentclass[letterpaper]{article} 
\usepackage{aaai2026}  
\usepackage{times}  
\usepackage{helvet}  
\usepackage{courier}  
\usepackage[hyphens]{url}  
\usepackage{graphicx} 
\usepackage{natbib}  
\usepackage{caption} 
\usepackage{algorithm}
\usepackage{algorithmic}
\usepackage{amsfonts}
\usepackage{booktabs}
\usepackage{multirow}
\usepackage{makecell} 
\usepackage{xcolor}
\usepackage{placeins}

\usepackage{newfloat}
\usepackage{listings}
\DeclareCaptionStyle{ruled}{labelfont=normalfont,labelsep=colon,strut=off} 
\floatstyle{ruled}
\newfloat{listing}{tb}{lst}{}
\floatname{listing}{Listing}
\usepackage{amsmath}
\usepackage{multirow}
\usepackage{color}
\usepackage{amssymb}
\usepackage{bibunits}
\usepackage[capitalize]{cleveref}

\title{\LARGE \bf
\ourmodel: 
One-Reference 6D Pose Estimation of Novel Objects via \\ Autoregressive Coordinate Map Generation}
\author{
    Dexin Zuo\textsuperscript{\rm 1}, Ang Li\textsuperscript{\rm 1}, Wei Wang\textsuperscript{\rm 2}\thanks{indicates corresponding authors.}, Wenxian Yu\textsuperscript{\rm 1}, Danping Zou\textsuperscript{\rm 1}\footnotemark[1]
}
\affiliations{
    \textsuperscript{\rm 1}Shanghai Jiao Tong University\\
    \textsuperscript{\rm 2}Corporate Research Center, State Key Laboratory of High-end Heavy-load Robots, Midea Group.\\
    
    \{dexin95, liang\_sjtu, wxyu, dpzou\}@sjtu.edu.cn, wangwei232@midea.com
}

\begin{document}

\maketitle

\begin{abstract}
Object 6D pose estimation, a crucial task for robotics and augmented reality applications, becomes particularly challenging when dealing with novel objects whose 3D models are not readily available. To reduce dependency on 3D models, recent studies have explored one-reference-based pose estimation, which requires only a single reference view instead of a complete 3D model. However, existing methods that rely on real-valued coordinate regression suffer from limited global consistency due to the local nature of convolutional architectures and face challenges in symmetric or occluded scenarios owing to a lack of uncertainty modeling. We present CoordAR, a novel autoregressive framework for one-reference 6D pose estimation of unseen objects. CoordAR formulates 3D-3D correspondences between the reference and query views as a map of discrete tokens, which is obtained in an autoregressive and probabilistic manner. To enable accurate correspondence regression, CoordAR introduces 1) a novel coordinate map tokenization that enables probabilistic prediction over discretized 3D space; 2) a modality-decoupled encoding strategy that separately encodes RGB appearance and coordinate cues; and 3) an autoregressive transformer decoder conditioned on both position-aligned query features and the partially generated token sequence. With these novel mechanisms, CoordAR significantly outperforms existing methods on multiple benchmarks and demonstrates strong robustness to symmetry, occlusion, and other challenges in real-world tests.
\end{abstract}

\begin{links}
    \link{Project Page}{https://sjtu-visys-team.github.io/CoordAR}
\end{links}

\section{Introduction}
\label{sec:introduction}
Object 6-DoF (Degrees of Freedom) pose estimation, which recovers the rotation and translation of a rigid object from observations, is a fundamental task in computer vision and robotics, with extensive applications in augmented reality, robotic manipulation, and industrial automation.  Despite its importance, real-world deployment remains challenging due to factors such as texture-less object surfaces, occlusion, and lighting variations.
 
 Learning-based approaches have made significant progress but often rely on known 3D models during training or inference and struggle to generalize to novel objects.  For instance, instance-level methods \cite{su2022zebrapose,liu2025gdrnpp} train a dedicated network per object using only synthetic data, achieving strong performance in the real world; however, they are costly and unflexible when it comes to novel objects. Category-level methods \cite{wang2019normalized,cai2024ov9d} improve generalization across intra-class variation but still struggle with out-of-distribution objects. 

 An alternative is to learn correspondences between image observations and a given 3D model \cite{nguyen2024gigapose,caraffa2024freeze}, enabling zero-shot pose estimation for novel objects. However, these methods typically assume access to textured CAD models at inference, which is usually an unrealistic assumption in many real-world scenarios involving unknown objects. 

\begin{figure}[!tpp]
    \centering
    \includegraphics[width=\linewidth]{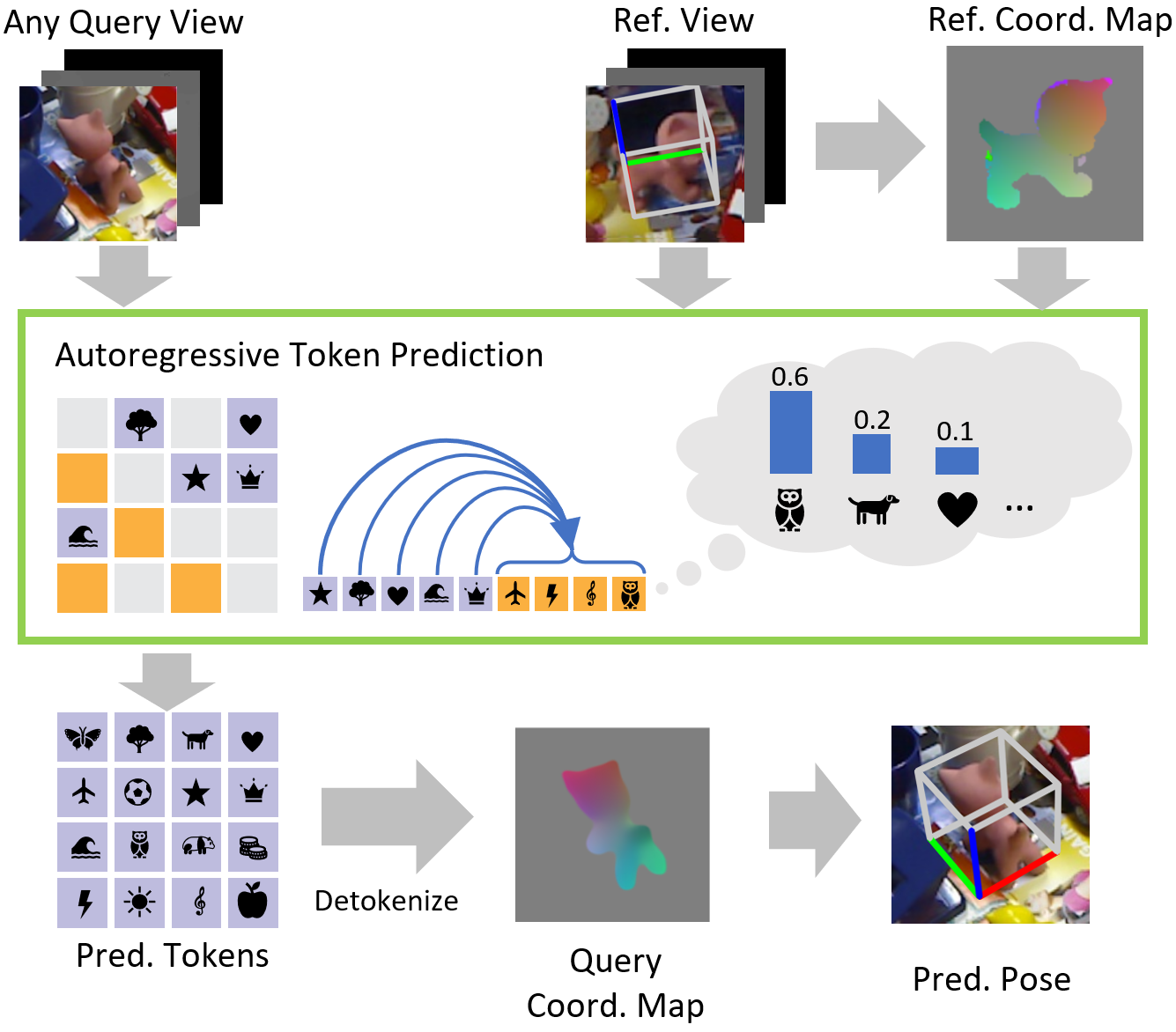}
    \caption{Given a reference view with known pose and depth-derived coordinate map, CoordAR  predicts the corresponding coordinates in the query view and subsequently obtain the relative pose from the correspondences provided by the coordinate maps. Instead of inferring continuous coordinate values in parallel, our model autoregressively operate on patch-level token space with a pretrained  tokenizer. }
    \label{fig:cover}
\end{figure}
To overcome this limitation, recent studies have turned to \emph{one-reference} methods, a promising paradigm that estimates the pose of a novel object using only a single reference view. Early methods \cite{corsetti2024open, fan2024pope,zhang2022relpose} relied on sparse correspondences between reference and target views; however, they struggled with texture-less surfaces, occlusions, and large viewpoint changes. More recently, One2Any~\cite{liu2025one2any} improves robustness by regressing coordinate maps as dense correspondences. However, it uses a convolutional decoder for real-valued coordinate regression, which introduces two key limitations.

Firstly, the limited receptive fields of the convolutional decoder restrict their ability to capture long-range dependencies, leading to inconsistent global reasoning in complex scenes. Secondly, training a regression model directly using continuous, real-valued coordinates fails to handle the inherent ambiguity arising from object symmetries and occlusions. Specifically, for symmetrical objects (e.g., cylinders, cubes), direct coordinate regression forces the network to reconcile multiple valid ground truths, leading to a wrong averaged result \cite{hodan2020epos}; for occluded objects, the model lacks an explicit mechanism to represent uncertainty in unobserved regions. These issues collectively reduce robustness in real-world applications.

In this paper, we propose \ourmodel, a novel framework for one-reference 6D pose estimation. Our model generates 3D-3D dense correspondences, represented by tokenized coordinate maps, conditioned on both the reference and query views. Based on the generated correspondences, the object pose can be calculated efficiently using the Umeyama algorithm~\cite{umeyama1991least}. Our model consists of three main stages: an encoding stage, where the reference RGB image and its coordinate map are encoded separately; a subsequent feature fusion stage; and a decoding stage, where the tokens are autoregressively decoded, conditioned on embeddings from both the reference and query views.

Our approach introduces three key innovations: (1) replacing traditional continuous coordinate regression with a probability prediction on a discretized space of the coordinate map, (2) introducing a modality-decoupled encoding strategy, where the RGB images and the reference coordinate map are encoded separately to obtain better performance and flexibility, and (3) designing a network that autoregressively generates the coordinate map, with the training objective formulated as predicting the conditional probability distribution of the next-set-of tokens given the query view, reference view, and the previously generated token sequence. Our method can accurately estimate 6D poses for novel objects in complex real-world scenarios using only a single reference view. Our contributions are : 
\begin{itemize}
    \item  To the best of our knowledge, we are the first to introduce autoregressive coordinate map generation for 6D pose estimation of novel objects. We further demonstrate the superiority of autoregressive generation compared to parallel real-valued regression. 
    \item We propose modality-decoupled encoders and transformer-style fusion blocks, integrating them into the framework, which effectively fuses information from both the query and reference views.
    \item We achieve state-of-the-art performance across multiple benchmark datasets, significantly outperforming existing one-reference methods.
\end{itemize}

\section{Related Works}
\paragraph{Model-based Methods}
Model-based object pose estimation methods leverage 3D models of target objects as prior knowledge. Existing model-based approaches can be broadly categorized into instance-level methods, category-level methods, and category-agnostic methods. Instance-level methods~\cite{xiang2017posecnn, Wang_2021_GDRN, li2019cdpn} operate on a closed set of known 3D models during training, which inherently restricts their application to previously seen objects.  While category-level methods~\cite{Wang_2019_CVPR, cai2024ov9d, chen2020learning, chen2021sgpa} demonstrate improved generalization to novel objects within trained categories, they remain constrained by their predefined taxonomic boundaries. Recently, some category-agnostic methods~\cite{labbe2022megapose,caraffa2024freeze,nguyen2024gigapose} estimate the relative pose based on the rendered anchor views of the 3D model. Despite their strong performance, reliance on CAD models limits their application to unseen real-world scenarios, where a novel object usually lacks a corresponding CAD model.
\paragraph{Model-free Methods}
To address scenarios where  3D models are unavailable, some methods~\cite{sun2022onepose, he2022onepose++, wen2024foundationpose, liu2022gen6d} first reconstruct the 3D model from multiple views with known poses and then estimate poses by comparing them with the images rendered from the reconstructed model. However, these methods rely on a sufficient number of views to build a high-quality 3D model, and performance drops significantly when views are sparse. Other methods directly compare the query view with sparse reference views. For example, FS6D~\cite{he2022fs6d} establishes 3D-3D correspondence through prototype matching between the query view and the reference views. 
Recently, the one-reference pose estimation problem has gained attention, where only a single reference view of the object is available. This setting poses significant challenges, primarily due to the wide diversity of object appearances, severe occlusions in both the reference and query views, and substantial viewpoint variations. One2Any~\cite{liu2025one2any} regresses a coordinate map that encodes the relative pose between the query view and a single reference view. However, it trains a convolutional decoder with a regression objective for continuous-valued target coordinate maps, which consequently struggles in challenging scenarios involving symmetries or heavy occlusion, often producing inaccurate coordinate predictions.

In this work, we explore autoregressive models~\cite{jiang2024survey,xiong2024autoregressive} to improve coordinate map regression, leveraging tokenization strategies that have demonstrated strong performance in tasks such as image generation~\cite{esser2021taming}, video generation~\cite{yu2023magvit} and embodied AI~\cite{micheli2022transformers}. Specifically, we introduce 1) a novel coordinate map tokenization scheme enabling probabilistic prediction over discretized 3D space, 2) a modality-decoupled encoding strategy that separately models RGB appearance and coordinate cues, and 3) an autoregressive transformer decoder conditioned on pixel-aligned query features and the partially generated coordinate sequence. Together, these novel mechanisms lead to significant improvements over state-of-the-art methods.

\begin{figure*}[!htbp]
    \centering
    \includegraphics[width=\linewidth]{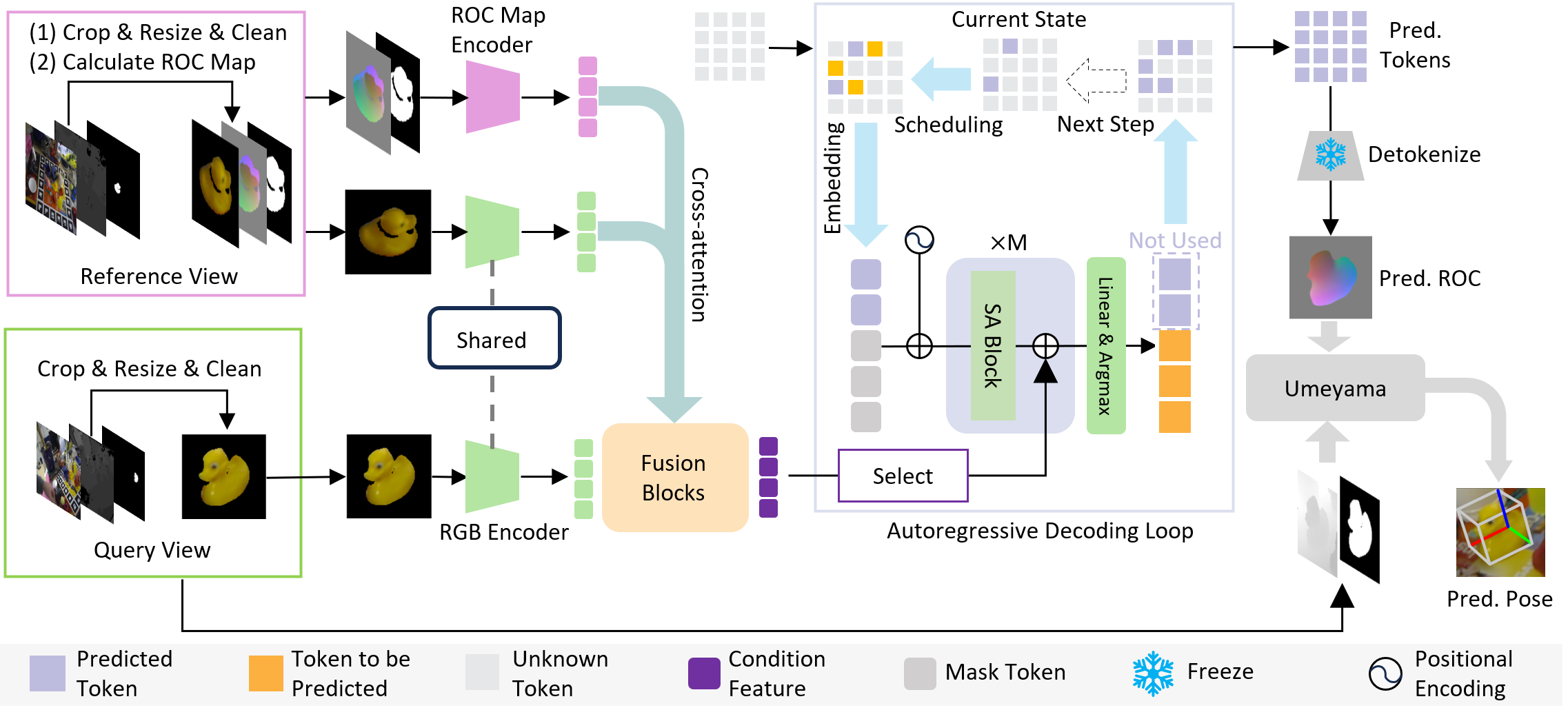}
    \caption{Overview of CoordAR framework. 
      Images from both query view and reference view are cleaned by masks to remove the interference from background and occlusion. The inputs are encoded by modality-decoupled encoders, where encoders are shared among same modality. The reference features are then integrated with the query features to form the condition feature. Subsequently, the decoder, which consists of several self-attention (SA) blocks, autoregressively decode new tokens with learned mask tokens as input. Finally, all tokens are detokenized and combined with the query depth to compute the pose.}
    \label{fig:main}
\end{figure*}

\section{Problem Statement}
Estimating the 6D pose of an object is a challenging yet practically valuable task, especially in scenarios where full 3D models are unavailable or object instances appear in dynamic or unstructured environments. Unlike traditional pose estimation settings that rely on known 3D CAD models or extensive multi-view observations, we focus on a one-reference setting, where the model must infer the pose of a novel object using only one annotated view as prior knowledge. To achieve this, we aim to estimate the relative transformation $\mathbf{T}_{RQ} \in SE(3)$ that transforms points from the query view to the reference view using the following inputs:

\begin{itemize}
    \item A reference RGB-D image $\mathcal{I}_R = (\mathcal{C}_R, \mathcal{D}_R)$ 
    where $\mathcal{C}_R$ denotes the color image and $\mathcal{D}_R$ is the depth map.
    \item A query RGB-D image $\mathcal{I}_Q = (\mathcal{C}_Q, \mathcal{D}_Q)$ from an unknown viewpoint.
    
    \item The object's binary mask $\mathcal{M}_R$ in the reference image and $\mathcal{M}_Q$ in the query image.
    
\end{itemize}
For evaluation, the absolute pose of the reference view $\mathbf{T}_{RO}$ is assumed to be known, allowing us to derive the absolute pose of the query view $\mathbf{T}_{QO}$.

\paragraph{Reference Object Coordinates (ROC) Map}
In our method, the relative object pose between the reference and query views is represented by Reference Object Coordinates (ROC) maps, an effective representation introduced by~\cite{liu2025one2any}. The ROC map of the reference view $\mathbf{X}^{R} \in \mathbb{R}^{H\times W\times 3}$ is obtained by backprojecting the depth within the reference mask and applying normalization:
\begin{equation}
    \mathbf{X}^R = \mathbf{S}\mathbf{\Pi}^{-1}(\mathcal{D}_{R})[\mathcal{M}_R=1],
    \label{eq:XR}
\end{equation}
where $\mathbf{\Pi}^{-1}(\cdot)$ denotes the backprojection operator, and $\mathbf{S} \in \mathbb{R}^{4 \times 4}$ is a normalization matrix that centers and scales the object point cloud. The normalization is derived from the inputs; further details are provided in the appendix. Likewise, the ROC map of the query view $\mathbf{X}^{Q} \in \mathbb{R}^{H\times W\times 3}$ is calculated as:
\begin{equation}
    \mathbf{X}^Q = \mathbf{S}\mathbf{T}_{RQ}\mathbf{\Pi}^{-1}(\mathcal{D}_{Q})[\mathcal{M}_Q=1],
    \label{eq:xQ}
\end{equation}
where $\mathbf{T}_{RQ}$ is the relative transformation from the query to the reference view. Both $\mathbf{X}^R$ and $\mathbf{X}^Q$ represent 3D points in the reference object frame, thereby providing pixel-wise 3D–3D correspondences between the query and reference images. Since $\mathbf{X}^R$ is known in advance, the task of 6D object pose estimation reduces to estimating the ROC map $\mathbf{X}^Q$ of the query image.

\section{The Proposed Method}
We introduce \ourmodel, a neural network for one-reference 6D object pose estimation. The overview of our method is shown in Fig.~\ref{fig:main}. Our network consists of three major stages: a modality-decoupled encoding stage, a subsequent fusion stage, and finally an autoregressive decoding stage, which we detail in the following sections.  The output of our network is a pixel-aligned ROC map $\mathbf{\hat{X}}^{Q} \in \mathbb{R}^{H \times W \times 3}$ that directly corresponds to the object's coordinates in the reference image, as described in Eq. (\ref{eq:xQ}).
\begin{figure}[!b]
    \centering
    \includegraphics[width=0.95\linewidth]{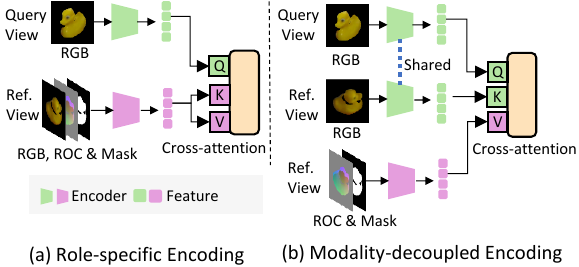}
    \caption{Different encoding schemes and cross-attention mechanisms in the fusion blocks. The modality-decoupled encoding improves architectural clarity and avoid affinity computation across modalities. To distinguish depth-absent regions from the background and invisible areas, we additionally concatenate the object mask to the input ROC map.}
    \label{fig:decouple}
\end{figure}

\paragraph{Modality-decoupled Encoding}
An encoding stage for both the query and the reference is a prerequisite for visual understanding. Existing work~\cite{liu2025one2any} leverages an encoder for the query and another for the reference, which we refer to as \textit{role-specific encoding}. To encode the reference information, they concatenate the RGB image and ROC map channel-wise as input, 
ignoring the distinctness in structural patterns between the two modalities.  In contrast to them, we assign separate encoders for different input modalities (RGB vs. ROC), allowing each encoder to specialize in its respective domain. As shown in Fig.~\ref{fig:decouple}, the modality-decoupled encoding we employ includes: 1) A \emph{shared} RGB encoder that processes both the reference image $\mathcal{C}_R$ and the query image $\mathcal{C}_Q$, and 2) Another encoder that handles the reference's ROC map $\mathbf{X}^{R}$. Details about each encoder are presented in the appendix.

\paragraph{Fusion Blocks} To condition token generation on reference-view cues, we introduce several stacked fusion blocks that integrate reference-view information with the query-view features.  Each fusion block has a similar structure to the decoder block in the transformer \cite{vaswani2017attention}, which primarily consists of a self-attention layer,  a following cross-attention layer, and a feedforward network (FFN). 
To accommodate modality-decoupled encoding, our cross-attention layer in the fusion blocks computes the affinity between encoded features within the same modality to mitigate the RGB-ROC domain gap, as demonstrated on the right side of Fig. \ref{fig:decouple}. This decoupling improves both architectural clarity and performance, as validated by our ablation studies. More details about the fusion blocks can be found in the appendix. Finally, the output features of the fusion blocks are considered the condition features for token generation. Specifically, when decoding a token at a certain position, the position-aligned condition feature is selected and added to the intermediate output of the decoder.

\paragraph{Autoregressive ROC Map Generation}
Instead of directly regressing the ROC map, our network first generates patch-level tokens and then detokenizes them to obtain the pixel-level ROC map. To this end, we adopt a VQ-VAE~\cite{van2017neural} as our ROC map tokenizer. The VQ-VAE first encodes the ROC map into latent vectors, then quantizes them by replacing each vector with its nearest neighbor in a pre-trained codebook $\mathcal{V}$, yielding a discrete token sequence $\{ s_1,\dots,s_{h \cdot w} \}$ where $s_* \in \mathcal{V}$. With the introduction of the VQ-VAE, the ROC map can be obtained indirectly by predicting discrete tokens, where a categorical distribution over $\mathcal{V}$ can be established at each patch. Unlike One2Any \cite{liu2025one2any}, where coordinates are generated in parallel, our decoder explicitly learns the dependencies between coordinates, which we find to be critical in our experiments. Mathematically, the distribution of tokens is represented as a masked autoregressive model \cite{li2024autoregressive} with the query and reference images as conditions:
\begin{equation}
    p(s)=\prod\limits_{k=1}^{K}p(S_{k}|S_{<k}, C_\mathbf{F}),
\end{equation}
where $S_{k}=\big\{s_i, s_{i+1}, \dots ,s_j\big\}$ are the tokens generated at the $k$-th step, and $s = \bigcup_{k=1}^{K} S_{k}$. Here $C_\mathbf{F}$ is the position-aligned condition feature, which is adapted from the feature after the fusion blocks.  The training objective of the autoregressive decoder is to minimize the negative log-likelihood loss:
\begin{equation}
    \mathcal{L_\text{AR}}=-\sum\limits_{k}^{K}\big[\log(p(S_k|S_{<k}, C_\mathbf{F}))\big].
\end{equation}
During inference, the previously generated tokens and position-aligned condition features both serve as conditioning information for predicting subsequent tokens. After all tokens have been generated, we leverage the decoder of the tokenizer to detokenize the tokens, producing an estimated ROC map $\mathbf{\hat{X}}^Q$. More details about the autoregressive decoder can be found in the appendix.

\paragraph{Recovering Object Pose from ROC Map}
As described in Eq.~(\ref{eq:xQ}), given the estimated ROC map $\hat{\mathbf{X}}^{Q}$, we recover the predicted 3D object points in the reference camera frame by applying the inverse of the normalization matrix:
\begin{equation}
    \mathbf{\hat{P}}_R^{Q} = \mathbf{S}^{-1} \mathbf{\hat{X}}^{Q},
\end{equation}
where $\mathbf{S}$ is the normalization matrix computed from the reference object points, as defined in Eq.~(\ref{eq:XR}).
To obtain the observed 3D points in the query camera frame, we backproject the depth map $\mathcal{D}_Q$ within the query mask:

\begin{equation}
    \mathbf{P}_Q^{Q} = \mathbf{\Pi}^{-1}(\mathcal{D}_Q)[\mathcal{M}_{Q}=1].
\end{equation}
where $\mathbf{\Pi}^{-1}(\cdot)$ is the camera backprojection operator.
Since $\mathbf{\hat{P}}_{R}^{Q}$ and $\mathbf{P}_{Q}^{Q}$ are pixel-aligned, we estimate the relative pose $\mathbf{T}_{RQ}$ using the Umeyama algorithm \cite{umeyama1991least},  which computes the optimal rigid transformation in a least-squares sense:
\begin{equation}
    \mathbf{\hat{T}}_{RQ} = \operatorname{Umeyama}(\mathbf{\hat{P}}_{R}^{Q}, \mathbf{P}_{Q}^{Q})
\end{equation}
Finally, given the object pose in the reference view $\mathbf{T}_{RO}$, the object pose in the query frame is obtained as $\mathbf{T}_{QO} = \hat{\mathbf{T}}^{-1}_{RQ}\mathbf{T}_{RO}$.

\section{Experiments}

\paragraph{Benchmark Datasets} To evaluate our method under various real-world scenarios, we consider four datasets: Real275~\cite{wang2019normalized}, Toyota-Light~\cite{hodan2018bop}, LINEMOD~\cite{hinterstoisser2011multimodal} and YCB-V~\cite{xiang2017posecnn}. These datasets encompass common challenges in 6D pose estimation, including illumination changes, occlusion, and significant variations in objects (geometric properties, materials, and textures), enabling a comprehensive evaluation of the algorithm.
\paragraph{Training Datasets} Consistent with the previous work \cite{liu2025one2any}, we train our models on the FoundationPose dataset~\cite{wen2024foundationpose} and a subset of the OO3D-9D~\cite{cai2024ov9d} dataset. See the appendix for more details.
\paragraph{Evaluation Metrics} To follow the baseline protocols for each setup, we evaluate pose estimation performance using the following metrics:
\begin{itemize}
    \item Recall of the ADD(-S) error, which is within 0.1 of the object diameter, as used in \cite{he2022fs6d,corsetti2024open}, shot for ADD(-S).
    \item Area under the curve (AUC) of ADD and ADD-S~\cite{xiang2017posecnn}.
    \item Average Recall of MSSD, MSPD, and VSD metrics defined in the BOP challenge~\cite{hodan2018bop}, shot for AR. 
\end{itemize}

\begin{table}[!b]
\centering
\fontsize{9pt}{10.8pt}\selectfont
\setlength{\tabcolsep}{1.2mm}
\begin{tabular}{c|c|cc|cc}
\toprule
\multicolumn{1}{c|}{\multirow{2}{*}{\textbf{Methods}}} & \multicolumn{1}{c|}{\multirow{2}{*}{\textbf{Modality}}} &  \multicolumn{2}{c|}{\textbf{Real275}} &\multicolumn{2}{c}{\textbf{Toyota-Light}} \\
& & AR & ADD(-S) & AR & ADD(-S) \\
\hline
LatentFusion  & RGB & 22.6 & 9.6 & 28.2 & 10.2 \\
ObjectMatch     & RGBD & 26.0 & 13.4 & 9.8 & 5.4 \\
Oryon        & RGBD & 46.5 & 34.9 & 34.1 & 22.9 \\
Any6D & RGBD & 51.0 & -- &43.3& --\\
One2Any    & RGBD & 54.9 & 41.0 & 42.0 & 34.6 \\
\hline
\textbf{\ourmodel} & RGBD & \textbf{71.0} & \textbf{82.2} & \textbf{62.5} & \textbf{82.6} \\
\bottomrule
\end{tabular} 
\caption{Comparison of 6D pose estimation methods on Real275~\cite{wang2019normalized} and Toyota-Light~\cite{hodan2018bop} datasets using AR and ADD(-S) metrics. Methods were evaluated on 2K reference-query image pairs. }
\label{tab:comparison}
\end{table}

\begin{figure}[h]
    \centering
    \includegraphics[width=0.95\linewidth]{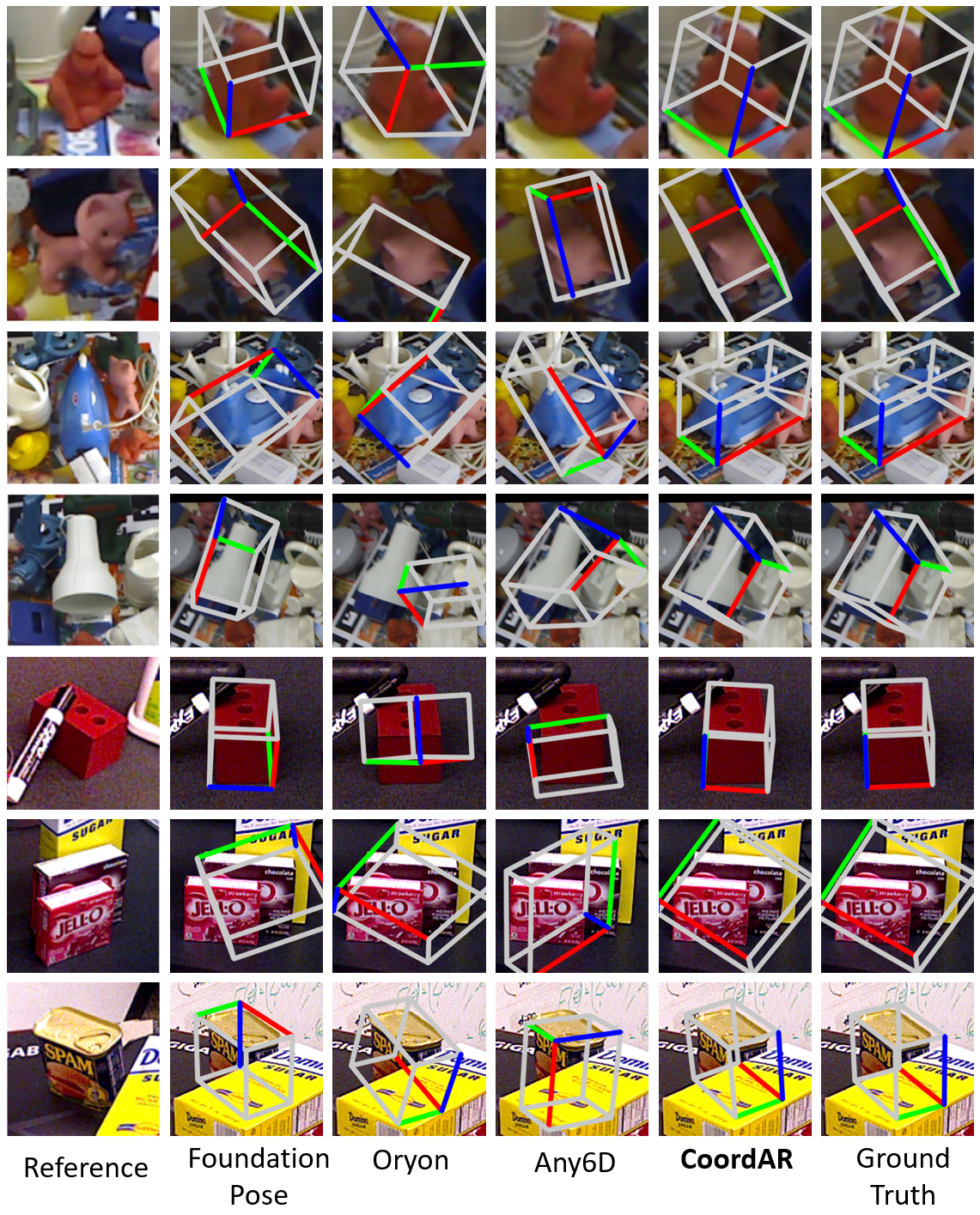}
    \caption{Visual comparison on LINEMOD and YCB-V datasets. We present results from several state-of-the-art methods alongside our method (\textbf{CoordAR}). Ground-truth poses are visualized in the last column, represented by bounding boxes with three distinctly colored edges.}
    \label{fig:qualitative}
\end{figure}

\begin{table*}[!htbp]
    \centering
    \fontsize{9pt}{10.8pt}\selectfont
   \setlength{\tabcolsep}{1mm}
    \begin{tabular}{c| c c | c c | c c || c c | c c | c c| c c}
    \hline
        \textbf{Methods}  &\multicolumn{2}{c|}{\textbf{Predator}} &\multicolumn{2}{c|}{\textbf{FS6D}} & \multicolumn{2}{c||}{\textbf{FoundationPose}} &  \multicolumn{2}{c|}{\textbf{FoundationPose}} &  \multicolumn{2}{c|}{\textbf{NOPE} } & \multicolumn{2}{c|}{\textbf{One2Any}} &\multicolumn{2}{c}{\textbf{\ourmodel}} \\
        \textbf{Ref. Images} & \multicolumn{2}{c|}{16} &  \multicolumn{2}{c|}{16} & \multicolumn{2}{c||}{16 - CAD}  & \multicolumn{2}{c|}{1 - CAD} & \multicolumn{2}{c|}{1 + GT trans} & \multicolumn{2}{c|}{1} & \multicolumn{2}{c}{1} \\
        \hline
         metrics of AUC & ADD & ADD-S &ADD & ADD-S & ADD & ADD-S & ADD & ADD-S & ADD & ADD-S & ADD & ADD-S & ADD & ADD-S\\
        \hline
can &  29.23 & 73.6  & 50.0 & 91.9 & \textbf{85.2} & \textbf{97.2} & \textbf{81.5} & 90.8 & 32.9 & \textbf{95.6} & 75.5 & 86.8 & 60.9 & 93.0 \\
box  & 21.33 & 62.58 & 45.0 & 93.1 & \textbf{94.4} & \textbf{98.0} & 81.2 & 91.1  & 20.3 & 85.3 & 90.2 & 78.7& \textbf{92.5} & \textbf{98.5} \\ 
bottle & 23.62 & 73.1& 39.1 & 87.7 & \textbf{90.5} & \textbf{97.0} & 73.3 & 90.0  & 26.7 & 89.3 & \textbf{90.7} & 93.0 & 87.5 & \textbf{99.7} \\ 
block & 22.75 & 74.85 &  36.8 & 95.2 & \textbf{94.1} & \textbf{97.8} & 36.8 & 96.7  &  35.7 & 95.0 & 84.9 & 91.1 & 84.6 & \textbf{97.3} \\
others & 24.1 & 71.72 & 40.2 & 83.2 & \textbf{94.9} & \textbf{97.5} & 40.2 & 93.4 & 32.2 & 86.4 & \textbf{81.3} & \textbf{95.2} & 75.4 & 93.1\\
\hline
mean & 24.3 & 71.0  & 42.1 & 88.4 & \textbf{91.5} & \textbf{97.4} & 76.1 &90.4 & 25.1 & 86.0 & \textbf{80.6} & 90.3 & 78.5 & \textbf{95.5}\\
         \hline
    \end{tabular}
    \caption{Performance on occluded YCB-V \cite{xiang2017posecnn} dataset.
    The Predator~\cite{huang2021predator}, originally proposed for point cloud registration, is additionally provided for reference. The methods are evaluated by AUC of ADD and AUC of ADD-S metrics. The baseline results are adopted from One2Any~\cite{liu2025one2any} and reproduced by their released model, where objects are categorized into five groups. Results on each object can be found in the appendix.}
    \label{tab:ycbv_data}
\end{table*}

\begin{table*}
\centering
    \fontsize{9pt}{10.8pt}\selectfont
    \setlength{\tabcolsep}{0.5mm}
\begin{tabular}{l|c c| c c c c c c c c c c c c c|c}
\hline
\textbf{Methods} & \textbf{Modality}  & \textbf{Ref. Images} & \textbf{ape} & \textbf{benchvise} & \textbf{cam} & \textbf{can} & \textbf{cat} &  \textbf{driller} & \textbf{duck} & \textbf{eggbox} & \textbf{glue} & \textbf{holepuncher} & \textbf{iron} & \textbf{lamp} & \textbf{phone} & \textbf{avg.} \\
\hline
OnePose & RGB & 200 & 11.8 & 92.6 & 88.1 & 77.2 & 47.9 & 74.5 & 34.2 & 71.3 & 37.5 & 54.9 & 89.2 & 87.6 & 60.6 & 63.6 \\
OnePose++  & RGB  & 200 & 31.2 & \textbf{97.3} & 88.0 & \textbf{89.8} & 70.4 & \textbf{92.5} & 42.3 & \textbf{99.7} & 48.0 & 69.7 & 97.4 & 97.8 & 76.0 & 76.9 \\
LatentFusion  & RGBD & 16 & \textbf{88.0} & 92.4 & 74.4 & 88.8 & 94.5 & 91.7 & 68.1 & 96.3 & 49.4 & 82.1 & 74.6 & \textbf{94.7} & 91.5 & 83.6 \\
FS6D + ICP & RGBD & 16 & 78.0 & 88.5 & \textbf{91.0} & 89.5 & \textbf{97.5} & 92.0 & \textbf{75.5} & 99.5 & \textbf{99.5} & \textbf{96.0} & 87.5 & \textbf{97.0} & \textbf{97.5} & \textbf{91.5} \\
\hline 
\hline
FoundationPose & RGBD & 1-CAD & 36.5 & 55.5 & \textbf{84.2} & 71.7 &  65.3 &16.3  & 49.8 & 42.6 & 64.8 & 52.7 & 20.7 & 15.8 & 51.7 & 48.3\\
NOPE & RGB & 1 + GT trans & 2.0 & 4.5 &  2.5 & 2.2 & 0.7 & 4.7& 0.5 &  \textbf{100.0} & 79.4 & 2.9  & 4.5 & 4.2 & 3.9 &16.3\\
Oryon &  RGBD & 1 & 1.2 & 1.3 & 3.9 & 0.8 & 12.7  & 8.5 & 0.8 & 63.2 & 18.4 & 1.6 & 0.6 &2.9 & 11.7 & 9.8 \\
One2Any & RGBD  & 1 & 33.1& 15.7 & 72.7 & 37.0 & 66.2  & 68.2 & 35.8 & \textbf{100.0} & \textbf{99.9} &  42.0 & 28.2 & 31.9 & 53.2 & 52.6 \\
\hline
\textbf{\ourmodel} & RGBD  & 1  & \textbf{45.6} & \textbf{76.9} & 70.7 & \textbf{77.3} & \textbf{88.1} & \textbf{96.5} & \textbf{50.2} & 97.0 & 99.8 & \textbf{67.5} & \textbf{52.7} & \textbf{91.4} & \textbf{61.2} & \textbf{75.0}\\
\hline
\end{tabular}
\caption{Performance on LINEMOD \cite{hinterstoisser2011multimodal} dataset with large view variations. We report the recall of ADD(-S) metric. Baseline results of taken from One2Any~\cite{liu2025one2any}.}.
    \label{tab:lm_data}
\end{table*}

\subsection{Results on Pose Estimation}
We primarily compare our method with model-free pose estimation approaches. For systematic comparison, our analysis includes model-free methods based on both single-view references and multi-view references. The single-view-based methods include Oryon~\cite{corsetti2024open}, ObjMatch~\cite{gumeli2023objectmatch}, NOPE~\cite{nguyen2024nope}, Any6D~\cite{lee2025any6d} and One2Any~\cite{liu2025one2any}; the multi-view-based methods include FoundationPose~\cite{wen2024foundationpose}, LatentFusion~\cite{park2020latentfusion} FS6D~\cite{he2022fs6d} OnePose~\cite{sun2022onepose} and OnePose++~\cite{he2022onepose++}. Baseline results are adopted from One2Any~\cite{liu2025one2any} with the assumption that ground-truth masks are available. Meanwhile, we follow the same evaluation protocols that they used. More specifically, on the LINEMOD~\cite{hinterstoisser2011multimodal} and YCB-V~\cite{xiang2017posecnn} datasets, the first view is chosen as the reference view for the entire test set. On the Real275~\cite{wang2019normalized} and Toyota-Light~\cite{hodan2018bop} datasets, 2K  reference-query image pairs are randomly sampled for evaluation.

\paragraph{Generalization to Real-world Novel Objects} We first evaluate our method on Real275~\cite{wang2019normalized} and Toyota-Light~\cite{hodan2018bop} for performance on real-world novel objects. Real275 contains 18 different real-world scenes comprising 42 unique objects across 6 categories, while Toyota-Light includes 21 rigid household objects under 5 different lighting conditions. As shown in Tab. \ref{tab:comparison}, our method surpasses the existing methods in both AR and ADD(-S) metrics, demonstrating excellent generalization to real-world novel objects. Qualitative results of the two datasets can be found in the appendix.

\paragraph{Occluded Scenes} The YCB-V~\cite{xiang2017posecnn} dataset contains numerous occluded scenes, including cases where even the first reference view is occluded. As displayed in rows 4 to 7 of Fig. \ref{fig:qualitative}, our method exhibits robustness when the query and reference are occluded. We also observe that our method performs well when the object frame is ill-defined (see row 6), suggesting that our method is independent of the definition of the canonical object frame. While we observe a slightly lower ADD AUC in Tab. \ref{tab:ycbv_data} compared to One2Any~\cite{liu2025one2any}, our method achieves a significantly higher ADD-S AUC than existing single-view methods. We attribute this discrepancy to a bias in the YCB-V evaluation protocol. Note that YCB-V contains texture-rich food containers that are geometrically symmetric but have different texture-based symmetry definitions during evaluation, such as the \textit{tomato\_soup\_can}. In such cases, ADD AUC can penalize predictions that are geometrically correct but differ in texture alignment, even though they achieve strong performance under the ADD-S metric (see the full YCB-V results in the appendix for details). Interestingly, when switching to the first frame of each scene as the reference, our method outperforms One2Any in both ADD AUC and ADD-S AUC, as demonstrated by the pose tracking results reported in the appendix.

\paragraph{Large View Variations} To evaluate robustness to large viewpoint variations, a comparison is conducted on the LINEMOD~\cite{hinterstoisser2011multimodal} dataset. This dataset features multiple texture-less objects, such as a toy ape and a hole-puncher. The images are captured by circling around each object, resulting in significant viewpoint variations. As displayed in Tab. \ref{tab:lm_data}, our approach achieves the highest performance across the majority of objects compared with single-view-based methods. Qualitative results are shown in rows 1 to 4 of Fig. \ref{fig:qualitative}. Notably, our method succeeds in estimating the pose of the top view (see row 2), the side view (see row 3), and even the back view (see row 1). While our method does not surpass the state-of-the-art multi-view approaches, it demonstrates overall superiority over OnePose~\cite{sun2022onepose} and achieves better performance on several objects  compared to OnePose++~\cite{he2022onepose++}.

\subsection{Ablation Studies}
\label{sec:ablation}
To justify the key design choices, we conduct ablation experiments on the LINEMOD dataset. Due
to computational limitations, models are trained with reduced iterations (see the appendix for more details).

\begin{figure}[!htbp]
    \centering
    \includegraphics[width=0.95\linewidth]{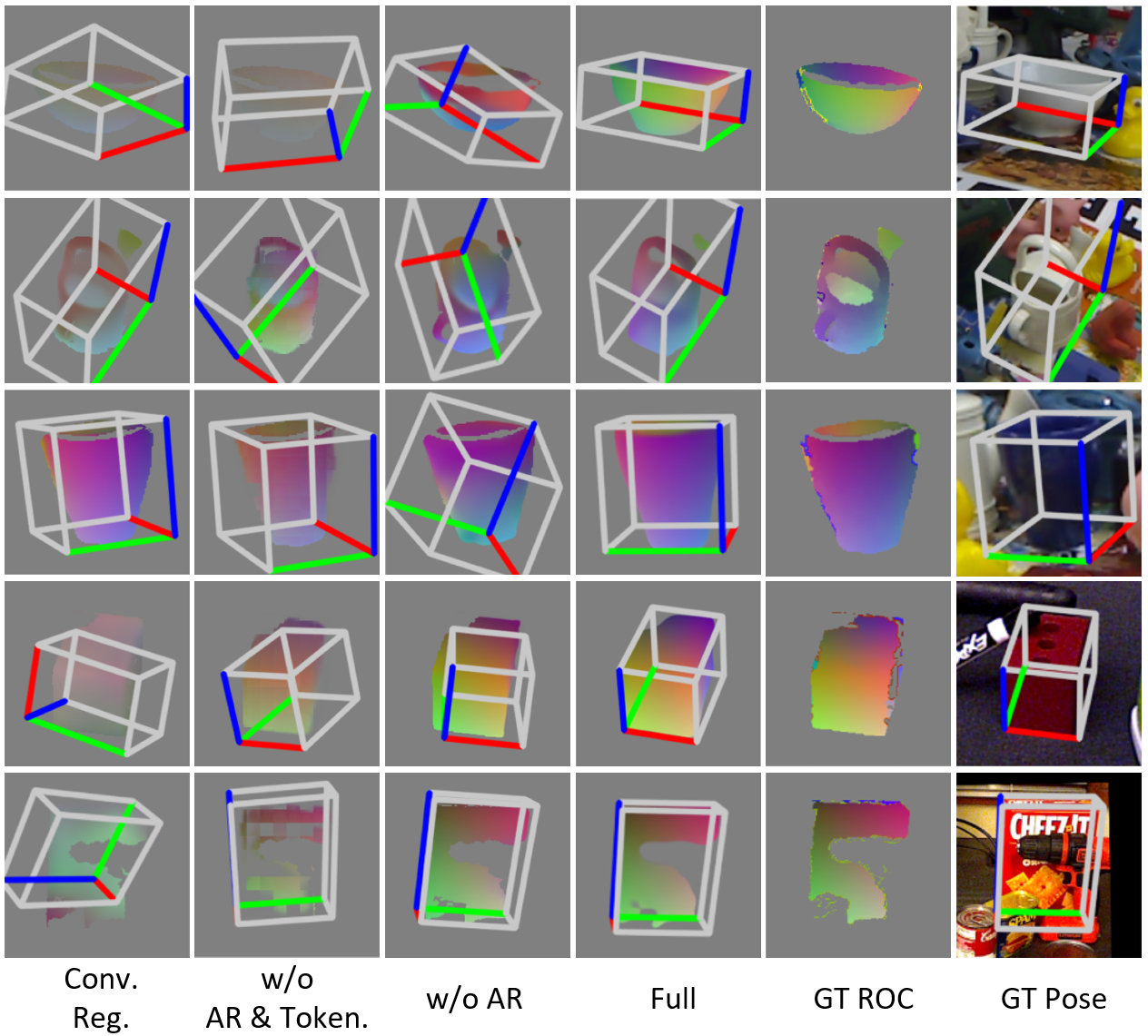}
    \caption{Qualitative Results of Ablation Study. We visualize the outputs of four decoder variants: (1) convolutional decoder, (2) full decoder without autoregressive tokenization, (3) full decoder without tokenization, and (4) full decoder. }
    \label{fig:ablation}
\end{figure}

\paragraph{Effect of Autoregressive Decoder}
We first study the overall effectiveness of the autoregressive decoder for ROC maps by comparing it against a convolutional regression decoder that has a similar architecture to the decoder in One2Any~\cite{liu2025one2any}. For a fair comparison, we reduce the number of parameters in our decoder to match those of the convolutional decoder. As shown in Tab. \ref{tab:ablation_study}, both metrics decrease after replacing the autoregressive decoder with the convolutional decoder. For further understanding, we keep the tokenization and disable the autoregression: 1) during both training and testing by training a model that predicts tokens in parallel, or 2) only during testing by generating all tokens in a single step. According to  Tab. \ref{tab:ablation_study}, disabling autoregression notably degrades performance. Fig. \ref{fig:ablation} provides a reasonable explanation for this result in columns 2 and 3, where non-autoregressive predictions exhibit  disrupted spatial coherence. For example, on the top-right of the \textit{can} (row 2, column 3), dark-purple values are incorrectly predicted  where light-green should appear. As demonstrated in Tab. \ref{tab:ablation_study}, disabling test-time autoregression reduces both ADD(-S) and AR by 5.6\%, decreasing ADD(-S) by 0.7\% when training is prolonged (see Tab. \ref{tab:inference_time}), suggesting that previously generated tokens can serve as an effective conditional context.

\begin{table}[!t]
    \centering
        \fontsize{9pt}{10.8pt}\selectfont
        \setlength{\tabcolsep}{0.4mm}
    \begin{tabular}{c  c | c | c | c }
    \hline
        \textbf{Component}  & \textbf{Variations} & \textbf{\#Params (B)}  & \textbf{ADD(-S)} & \textbf{AR}  \\
       \hline
        \multirow{2}{*}{\makecell[c]{ROC decoding}} & 
 convolutional & 0.28  & 70.9 & 59.7 \\
        & autoregressive & 0.28 & \textbf{73.1} & \textbf{61.6} \\
        \hline 
        \multirow{2}{*}{\makecell[c]{Autoregression}} & w/o& 0.37 & 60.7 & 52.1 \\
        & w/ & 0.37 & \textbf{73.6} & \textbf{61.9} \\
         \hline
              \multirow{2}{*}{\makecell[c]{Test-time AR}} 
        & w/o  & 0.37 & 68.0 & 56.3 \\
        & w/ & 0.37 &\textbf{73.6} & \textbf{61.9} \\
        \hline
        \multirow{2}{*}{\makecell[c]{Tokenization}} & w/o & 0.37 & 56.4 & 48.7 \\
        & w/ & 0.37 & \textbf{60.7} & \textbf{52.1} \\
        \hline
        \multirow{2}{*}{\makecell[c]{Encoding}} & role-specific  & 0.37 & 61.6 & 49.8 \\
         & modality-decoupled & 0.37 & \textbf{73.6} & \textbf{61.9} \\
        \hline
    \end{tabular}
    \caption{Ablation study on critical design choices. All evaluations are conducted on the full LINEMOD dataset using the AR and ADD(-S) metrics. Parameter counts (in billions, excluding the tokenizer) are provided for reference. }
    \label{tab:ablation_study}
\end{table}

\paragraph{Effect of Tokenization}
Afterward, we further remove the reliance on tokenization by regressing real-valued ROC maps from the features of the pre-final layer in the transformer decoder. As quantified in Tab. \ref{tab:ablation_study}, replacing  token prediction with real-value regression further decreases recall. This may be attributed to their handling of ambiguous cases. As depicted in Fig. \ref{fig:ablation}, this model produces ambiguous ROC maps when dealing with symmetry ambiguity on the \textit{bowl} (row 1, column 2) and occlusion ambiguity on the \textit{cup} (row 3, column 2). A similar failure also occurs in the convolutional-head variant (see results on the \textit{bowl} in row 1 and the \textit{box} in row 5). Notably, discrete-token-based models (columns 3–4) demonstrate improved performance with clearer coordinate maps, suggesting that probabilistic modeling plays a key role in resolving such ambiguities.

\paragraph{Effect of Modality-decoupled Encoding} To validate the effectiveness of the modality-decoupled encoding, we replace it with the role-specific encoding used in One2Any~\cite{liu2025one2any}. More specifically, we employ two DINOv2-structured encoders: one processing the query RGB image and another handling the channel-wise concatenation of the reference RGB image and its corresponding ROC map and mask. As shown in Tab. \ref{tab:ablation_study}, the performance degrades significantly after switching to role-specific encoding, indicating that allocating encoders by modality is critical for visual understanding.

\subsection{Runtime Analysis}
Our model supports trade-offs between accuracy and computational efficiency by adjusting the number of generation steps. Inference speed comparisons, along with their corresponding ADD(-S) on the LINEMOD dataset, are provided in Tab. \ref{tab:inference_time}. Impressively, our model can achieve near real-time speed (0.10 seconds per frame) with even a single step while maintaining comparable accuracy.

\begin{table}[!t]
    \centering
        \fontsize{9pt}{10.8pt}\selectfont
        \setlength{\tabcolsep}{0.4mm}
    \begin{tabular}{c| c| c | c | c }
         \toprule
        \textbf{Methods} & \textbf{Modality} & \textbf{GPU} &\textbf{Time (s)} & \textbf{ADD(-S)}  \\
       \hline
        FoundationPose & 1-CAD & RTX 4090 & 2.70 & 48.3 \\
        One2Any & RGBD & RTX4090 & \textbf{0.09} & 56.2 \\
        \hline
        Ours-64 steps & RGBD & RTX 4090 & 0.63 & \textbf{75.0} \\
        Ours-16 steps & RGBD & RTX 4090 & 0.25 & \textbf{75.0} \\
        Ours-4 steps & RGBD & RTX 4090 & 0.13 & 74.7 \\
        Ours-1 step & RGBD & RTX 4090 & 0.10 & 74.3 \\
         \hline
    \end{tabular}
    \caption{Inference time comparison. The runtimes of 
    FoundationPose~\cite{wen2024foundationpose} and One2Any~\cite{liu2025one2any} are taken from One2Any~\cite{liu2025one2any}.}
    \label{tab:inference_time}
\end{table}

\section{Conclusions}
In this paper, we propose the first autoregressive framework for one-reference 6D pose estimation of novel objects. By formulating correspondence prediction as an autoregressive probabilistic token decoding task and introducing modality-decoupled encoding for visual understanding, CoordAR achieves superior performance on standard benchmarks. Extensive experiments demonstrate significant improvements in handling symmetry, occlusion, and novel objects. This work establishes autoregressive coordinate modeling as a promising direction for robust 6D pose estimation.

\section{Acknowledgments}
This work was supported in part by the National Key R\&D
Program of China under Grant 2022YFB3903801, in part by the National Science Foundation of China under Grant 62073214, and in part by the Corporate Research Center, State Key Laboratory of High-end Heavy-load Robots, Midea Group (3D
Robot Vision Project).
We thank Kun Wang, Li Shen, Yuhui Ni, and Yikun Zeng for providing the reproduced results of baseline methods for qualitative comparison.

\FloatBarrier
\bibliography{aaai2026}

\clearpage
\newcommand{\isSuppMainFile}{}
\makeatletter
\@ifundefined{isSuppMainFile}{
  \newif\ifreproStandalone
  \reproStandalonetrue
}{
  \newif\ifreproStandalone
  \reproStandalonefalse
}
\makeatother

\ifreproStandalone
\documentclass[letterpaper]{article}
\usepackage[submission]{aaai2026}
\setlength{\pdfpagewidth}{8.5in}
\setlength{\pdfpageheight}{11in}
\usepackage{times}
\usepackage{helvet}
\usepackage{courier}
\usepackage{xcolor}
\usepackage{amsmath}
\usepackage{multirow}
\usepackage{color}
\usepackage{amssymb}
\usepackage{natbib}
\usepackage{url}
\usepackage{booktabs}
\usepackage{makecell} 
\usepackage{placeins}
\usepackage{graphicx}
\usepackage[capitalize]{cleveref}

\newcommand{\ourmodel}{CoordAR}

\frenchspacing

\begin{document}
\fi

\renewcommand{\thefigure}{A\arabic{figure}}  
\renewcommand{\theequation}{A\arabic{equation}} 
\renewcommand{\thetable}{A\arabic{table}} 
\setcounter{secnumdepth}{2}

\section*{Appendix}
In this appendix, we provide detailed information about data preparation, method implementation, and experimental results.

\section{Data Preparation}

\subsection{Data Preprocessing}
\subsubsection{Normalization Matrix} The normalization transformation is constructed by first moving the origin to the center of the object, denoted as $\mathbf{c}=[c_x, c_y, c_z]^{\mathrm{T}}$, and then applying a scale normalization according to the diameter of the object. The normalization matrix is calculated as:
\begin{equation}
    \mathbf{S}= \begin{bmatrix}
    1/d & 0 & 0 & 0 \\
    0 & 1/d & 0 & 0 \\
    0 & 0 & 1/d & 0 \\
    0 & 0 & 0 & 1
\end{bmatrix} \cdot \begin{bmatrix}
    1 & 0 & 0 & -c_x \\
    0 & 1 & 0 & -c_y \\
    0 & 0 & 1 & -c_z \\
    0 & 0 & 0 & 1
\end{bmatrix},
\end{equation}
where $d$ is the diameter of the object. For simplicity, our normalization uses the diameter as a single size parameter, which is different from that of One2Any~\cite{liu2025one2any}, where a size parameter is calculated for each axis.  Precisely estimating the size parameters of an object is challenging from a single reference depth image, particularly when the object is occluded by itself or by other objects. Moreover, the estimation is vulnerable to depth noise, which can introduce outliers. To overcome this, we estimate the diameter from the 2D query mask and the median depth. Specifically,  the diameter is approximated by:
\begin{equation}
    d=\frac{D_{median}\cdot \sqrt{w^2+h^2}}{(f_x+f_y)/2},
\end{equation}
where $\{w, h\}$ (in pixels) are the width and height of the visible area, respectively; $D_{median}$ is the median depth in the visible area, and $\{f_x, f_y\}$ are the camera intrinsic focal lengths (in pixels) along the 
x- and  y-axes, respectively.

\subsection{Training Data}

We apply data augmentations during training. The RGB images are augmented with random backgrounds using images from the PASCAL VOC dataset~\cite{everingham2010pascal}. The reference ROC maps are randomly corrupted, with parts masked out. 

\section{Implementation Details}

\subsection{The Image Encoders}
For RGB image encoding, we employ a DINOv2~\cite{oquab2023dinov2} backbone with trainable parameters, leveraging its powerful pretrained encoder while allowing for fine-tuning to adapt to pose estimation requirements. To ensure dimensional consistency between modalities, we encode ROC maps  using a ViT-B encoder~\cite{dosovitskiy2020image}.

\subsection{The Tokenizer}
To ensure compatibility with our RGB feature encoder's output scale, the tokenizer generates compact token maps at $1/f$ the input resolution ($f=16$ for our tokenizer). 
Formally, given an input ROC map $\mathbf{X}^{Q} \in \mathbb{R}^{H \times W \times 3}$, the encoder of the VQ-VAE $\mathcal{T}_{e}(\cdot) $ encodes $\mathbf{X}^{Q}$ to a continuous latent vector map:
    \begin{equation}
        \mathbf{z}_e = \mathcal{T}_{e}(\mathbf{X}^Q) \in \mathbb{R}^{h \times w \times d}, \quad \text{where} 
        \begin{cases}
            h = \lfloor H/f \rfloor, \\
            w = \lfloor W/f \rfloor
        \end{cases}.
    \end{equation}
Each latent vector is then quantized using a learned codebook $\mathcal{B} = \{\mathbf{e}_k\}_{k=1}^K \subset \mathbb{R}^d$:
    \begin{equation}
        \mathbf{z}_q^{i} = \mathop{\text{argmin}}_{\mathbf{e}_k \in \mathcal{B}} \|\mathbf{z}_e^{i} - \mathbf{e}_k\|_2
    \end{equation}
where $\mathbf{z}_e^{i},\mathbf{z}_q^{i}$ denotes the latent vectors at position $i \in \{1,\dots,h\cdot w\}$. The discrete tokens are the indices of the quantized vectors in the codebook: $\{ s_1,\dots,s_{h\cdot w} \}$.

We use the ground-truth $\mathbf{X}^{Q}$ sampled from the training set to train the tokenizer. The training objective of the tokenizer is as follows:
\begin{align}
\mathcal{L}_{\text{VQ-VAE}} = &
\left\| \mathbf{X}^{Q} - \mathcal{T}_{d}(\mathbf{z}_q) \right\|^2 
+ \left\| \text{sg}(\mathbf{z}_e) - \mathbf{z}_q \right\|^2 \\
& + \beta \left\| \mathbf{z}_e - \text{sg}(\mathbf{z}_q) \right\|^2, \nonumber
\end{align}
where $\mathbf{z}_q$ is the latent vector returned by the encoder of the tokenizer, $\mathbf{z}_e$ is the embedding vector found by nearest lookup in the codebook, and $\text{sg}(\cdot)$ is the stop gradient operator to prevent gradients from back-propagating through its argument during training.

\subsection{The Fusion Blocks}
\begin{figure*}[!htbp]
    \centering
    \includegraphics[width=0.9\linewidth]{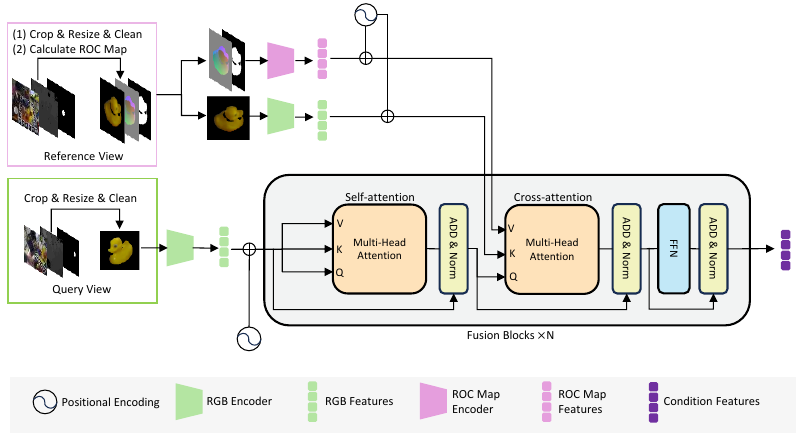}
    \caption{Architecture of the Fusion Blocks. The reference features and query features are integrated by the fusion blocks. The output condition feature is further used for token decoding.}
    \label{fig:fusion_blocks}
\end{figure*}
As illustrated in \cref{fig:fusion_blocks}, we employ N (N=8) fusion blocks to integrate information from both the query and reference. Each fusion block primarily consists of a self-attention layer, a cross-attention layer, a feed-forward network (FFN), and layer normalizations. To incorporate positional information, the features are added with positional encoding before being fed into the blocks. The final output is a set of conditional features used for decoding.

\subsection{The ROC Decoder}
\begin{figure*}
    \centering
    \includegraphics[width=0.9\linewidth]{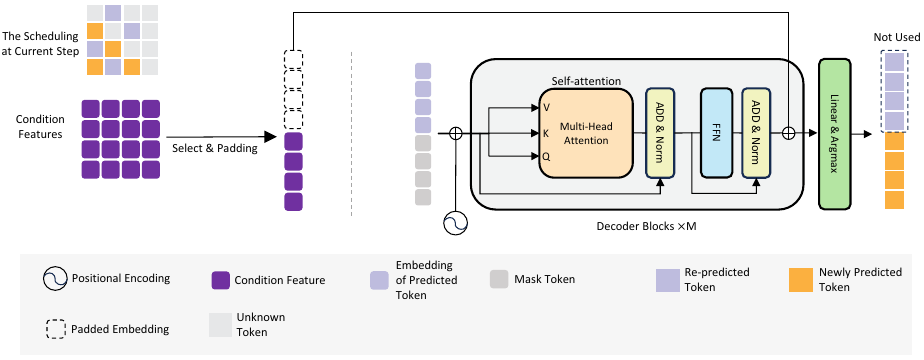}
    \caption{Architecture of the ROC Decoder. This component autoregressively decode tokens conditioned on the condition features. Condition features are padded for a consistent length with the input of the decoder.}
    \label{fig:roc_decoder}
\end{figure*}

As depicted in \cref{fig:roc_decoder}, the ROC decoder operates autoregressively at each generation step by processing both the sequence of previously generated tokens and the conditional features through multiple stacked decoding layers to predict the subsequent tokens. The embeddings for these generated tokens are retrieved by querying the tokenizer's codebook, while the conditional features aligned with the target token position are selectively incorporated into the final layer of each decoding block. 

\subsection{Training Details}
Training an ROC-map tokenizer is a prerequisite for our main network. The tokenizer is trained using ROC maps of the query-reference pairs sampled from the training set. The training of the tokenizer lasts for 200k iterations. After being trained, the tokenizer is frozen in the following steps. Subsequently, the coordinate map estimation network is trained for 400K iterations. We employ the Ranger~\cite{Ranger} optimizer with a OneCycle~\cite{smith2019super} learning rate scheduler.  We conduct all training on a server equipped with 4 NVIDIA RTX 4090 GPUs. The tokenizer training completes within several hours, while the main network requires approximately 5 days to converge. For both the tokenizer and the main network, the batch size is set to 64. Due to computational limitations, the number of training iterations for the ablation study is set to 200K, with a batch size of 32 for all models. 

\section{Experimental Results}

\subsection{Full Results on the YCB-V Dataset}
In Tab. \ref{tab:ycbv_data_full}, we present the performance of each object in the YCB-V dataset. 
As shown in the table, our method demonstrates an inferior ADD AUC compared to One2Any~\cite{liu2025one2any}, which we attribute to both the evaluation bias of the ADD AUC and the low texture overlap between views. First, the ADD AUC can produce a low score even when the estimated pose is geometrically correct. An example of a geometrically
correct pose is presented in Fig. A3.
Moreover, as shown in Tab. \ref{tab:ycbv_tracking_full}, no degradation in ADD AUC is observed in the YCB-V tracking benchmark, and our method outperforms One2Any. This is possibly because the reference changes from the first view of all scenes to the first frame of each scene, where similarly textured regions are more likely to appear in both the reference and query views.

\subsection{Inference with Multi-view References}
Our method can be further extended to a multi-view setup in which multi-view references and their poses are available. We generate a pose hypothesis from each reference and then adopt the pose selection introduced by One2Any~\cite{liu2025one2any} to select the best pose, where the reference masks are re-projected by the relative pose and the best pose is selected based on the mIoU between the re-projected reference mask and the query mask. As demonstrated in Tab. \ref{tab:supp_multiview}, our method surpasses One2Any and existing multi-view-based methods, except for FS6D~\cite{he2022fs6d}. Since the mIoU based selection does not guarantee the optimal selection~\cite{liu2025one2any}, we further  present the performance with optimal selection, where the pose of the  best view is selected. Our method demonstrates superior performance compared to other approaches with optimal view selection, suggesting a strong potential for further enhancement in the multi-view setting.

\begin{table*}[!htbp]
\centering
    \fontsize{9pt}{10.8pt}\selectfont
    \setlength{\tabcolsep}{0.5mm}
\begin{tabular}{l|c c| c c c c c c c c c c c c c|c}
\hline
\textbf{Methods} & \textbf{Modality}  & \textbf{Ref. Images} & \textbf{ape} & \textbf{benchvise} & \textbf{cam} & \textbf{can} & \textbf{cat} &  \textbf{driller} & \textbf{duck} & \textbf{eggbox} & \textbf{glue} & \textbf{holepuncher} & \textbf{iron} & \textbf{lamp} & \textbf{phone} & \textbf{avg.} \\
\hline
OnePose & RGB & 200 & 11.8 & 92.6 & 88.1 & 77.2 & 47.9 & 74.5 & 34.2 & 71.3 & 37.5 & 54.9 & 89.2 & 87.6 & 60.6 & 63.6 \\
OnePose++  & RGB  & 200 & 31.2 & 97.3 & 88.0 & 89.8 & 70.4 & 92.5 & 42.3 & 99.7 & 48.0 & 69.7 & 97.4 & 97.8 & 76.0 & 76.9 \\
LatentFusion  & RGBD & 16 & 88.0 & 92.4 & 74.4 & 88.8 & 94.5 & 91.7 & 68.1 & 96.3 & 49.4 & 82.1 & 74.6 & 94.7 & 91.5 & 83.6 \\
FS6D + ICP & RGBD & 16 & 78.0 & 88.5 & 91.0 & 89.5 & 97.5 & 92.0 & 75.5 & 99.5 & 99.5 & 96.0 & 87.5 & 97.0 & 97.5 & 91.5 \\
One2Any & RGBD  & 16-mIoU & 82.1 & 85.5 & 92.8 & 75.9 & 94.1 & 80.4 & 65.9 & \textbf{100.0} & 99.9 &  70.7 & 61.7 & 91.5 & 84.1 & 83.7  \\
One2Any & RGBD  & 16-best view & 84.8 & 98.3 & 98.8 & 95.2 & 95.9 & 93.3 & 76.2 & \textbf{100.0} & 99.9 & 92.9 & 95.1 & 94.4 & 93.9 & 93.8\\
\hline
\ourmodel & RGBD & 16-mIoU & 85.0 & 99.9 & 79.4 & 95.1 & 93.3 & 96.4 & 66.3 & 98.2 & 100.0 & 93.2 & 79.9 & 94.1 & 88.0 & 89.9 \\
\ourmodel & RGBD & 16-best view & \textbf{95.1} & \textbf{100.0} & \textbf{99.7} & \textbf{100.0} & \textbf{99.7} & \textbf{99.4} & \textbf{96.0} & 99.9 & \textbf{100.0} & \textbf{99.6} & \textbf{99.3} & \textbf{99.8} & \textbf{99.4}& \textbf{99.1}\\
\hline
\end{tabular}
\caption{Multiview performance on LINEMOD \cite{hinterstoisser2011multimodal}. Baseline results of taken from One2Any~\cite{liu2025one2any}.}.
    \label{tab:supp_multiview}
\end{table*}

\subsection{Ablation on Inference Configurations}
The autoregressive framework provides multiple prediction choices during inference. To investigate how these design choices affect pose accuracy, we conducted experiments on the LINEMOD dataset~\cite{hinterstoisser2011multimodal}.

\paragraph{Generation Steps} Our initial evaluation focused on model performance under varying generation steps. As shown in Tab. \ref{tab:inference_ablation}, our model achieves optimal pose accuracy at 64 steps, with gradually degrading performance as the step count decreases. This demonstrates that predicting new tokens conditioned on preceding tokens is a critical architectural consideration. Notably, even when reduced to a single step (step=1), our model maintains reasonable accuracy without catastrophic failure. This property enables deployment in latency-sensitive applications, such as pose tracking, which is discussed in the following sections.

\paragraph{Token Scheduler} In our implementation, the number of tokens to be predicted in each generation step is controlled by a token scheduler. To investigate the sensitivity to the type of scheduler, we apply a linear scheduler and a cosine scheduler individually at evaluation time. As displayed in Tab. \ref{tab:inference_ablation}, the cosine scheduler, which progressively increases generation speed after an initial cautious phase, results in measurable performance gains, empirically validating the need for deliberate token generation during high-ambiguity initial stages.

\paragraph{Generation Order} Generating images by random order or raster scan order is a common practice in autoregressive image-generation models~\cite{xiong2024autoregressive}. To investigate whether performance can be improved by a better global setting of generation order, we evaluate our method using random order and raster scan order. As presented in Tab. \ref{tab:inference_ablation}, random order and raster-scan order show comparable performance, suggesting that the overall performance is invariant to a global setting of order.

\noindent\textbf{Randomness in Inference.} In our method, we use $\operatorname{argmax}$ to select tokens from the predicted distribution deterministically. However, probabilistic token sampling from the distribution with a temperature parameter $\tau$ can also be an option. As displayed in Tab. \ref{tab:inference_ablation}, we observe a performance degradation after adding randomness to the inference procedure, indicating that deterministic token selection is preferable for maintaining pose estimation accuracy.


\begin{table}[!htbp]
    \centering
        \fontsize{9pt}{10.8pt}\selectfont
        \setlength{\tabcolsep}{0.4mm}
    \begin{tabular}{c  c | c | c | c }
        \hline
        \multirow{2}{*}{\textbf{Component}} & \multirow{2}{*}{\textbf{Variations}}  & \textbf{AUC}& \multirow{2}{*}{\textbf{ADD(-S)}} & \multirow{2}{*}{\textbf{AR}} \\
        &&\textbf{of ADD(-S)}&&\\
        \hline
        \multirow{4}{*}{\makecell[c]{Generation steps}} 
        & 1  & 73.5 & 68.0 & 56.3 \\
        & 4  & 74.2 & 69.5& 58.4 \\
        & 16  & \textbf{74.4} & 73.2 & 61.8 \\
        & 64 & 74.3 &\textbf{73.6} & \textbf{61.9} \\
        \hline
        \multirow{2}{*}{\makecell[c]{Token scheduler}} & linear  & 74.2& 72.4 & 60.4 \\
        & cosine & \textbf{74.3}& \textbf{73.6} & \textbf{61.9}\\
        \hline
        \multirow{2}{*}{Generation order} & raster scan &  \textbf{74.3} & \textbf{73.6} & \textbf{62.1} \\
        & random  & \textbf{74.3}& \textbf{73.6} & 61.9 \\
        \hline
        \multirow{3}{*}{Randomness} & $\tau=1.0$  & 72.1 & 69.4 &  57.6 \\ 
        & $\tau=0.5$& 72.6 & 70.6 & 58.2 \\
        & argmax  & \textbf{74.3} & \textbf{73.6} & \textbf{61.9} \\
        \hline
    \end{tabular}
    \caption{Performance with different inference configurations. The last row of each configuration is our default setting.}
    \label{tab:inference_ablation}
\end{table}

\begin{table*}[!htbp]
    \centering
    \footnotesize
       \setlength{\tabcolsep}{0.2mm}
    
    \begin{tabular}{c| c c |c c |c c || c c | c c | c c | c c}
    \toprule
        Methods  &\multicolumn{2}{c|}{\textbf{PREDATOR} } & \multicolumn{2}{c|}{\textbf{FS6D}} & \multicolumn{2}{c||}{\textbf{FoundationPose}} &  \multicolumn{2}{c|}{\textbf{FoundationPose}} &  \multicolumn{2}{c|}{\textbf{NOPE}} & \multicolumn{2}{c|}{\textbf{One2Any}}& \multicolumn{2}{c}{\textbf{\ourmodel}} \\
  
        Ref. Images & \multicolumn{2}{c|}{16} & \multicolumn{2}{c|}{16} & \multicolumn{2}{c||}{16 - CAD}  & \multicolumn{2}{c|}{1 - CAD} & \multicolumn{2}{c|}{1 + GT trans} & \multicolumn{2}{c|}{1} & \multicolumn{2}{c}{1} \\
        \hline
        metrics of AUC & ADD & ADD-S  & ADD & ADD-S  & ADD & ADD-S & ADD & ADD-S & ADD & ADD-S & ADD & ADD-S & ADD & ADD-S \\
        \hline
       002\_master\_chef\_can*  &17.4 & 73.0 & 36.8 & 92.6 & \textbf{91.3} & \textbf{96.9} &73.3 & 87.3 & 17.8 & \textbf{96.8} & \textbf{82.8} & 94.4 & 42.3 & \textbf{96.8}\\
003\_cracker\_box* & 8.3 & 41.7  & 24.5 & 83.9 & \textbf{96.2} & \textbf{97.5} & 72.2	& 92.0 & 2.8 & 83.0 & 74.9 & 82.6 & 72.4 & \textbf{94.3}\\
004\_sugar\_box &  15.3 & 53.7 & 43.9 & 95.1 & \textbf{87.2} & \textbf{97.5} &  87.1	&88.2  & 22.3 & 86.5 & 93.0 & 97.7 & \textbf{97.6} & \textbf{99.9}\\
005\_tomato\_soup\_can* & 44.4 & 81.2 & 54.2 & 93.0 & \textbf{93.3} & \textbf{97.6} & 92.3 &	95.2  & 48.4 & \textbf{95.9}  & 78.6 & 88.4 & 71.8 & 95.4 \\
006\_mustard\_bottle &  5.0 & 35.5 &  71.1 & 97.0 & \textbf{97.3} & \textbf{98.4} & 76.6 &	88.4 & 42.7 & 91.3 & 93.2 & \textbf{100.0} & 74.3 & 99.8\\
007\_tuna\_fish\_can* &  34.2 & 78.2 & 53.9 & 94.5 & \textbf{73.7} & \textbf{97.7} & 76.9 &	90.5 &33.3 & 97.0 & \textbf{80.8} & 86.6 & 65.1 & \textbf{98.3}\\
008\_pudding\_box &  24.2 & 73.5 & 79.6 & 94.9 & \textbf{97.0} & \textbf{98.5} & 77.8 &	91.7 & 20.9 & 84.4 & 72.1 & 73.4 & \textbf{100.0}&\textbf{100.0} \\
009\_gelatin\_box &  37.5 & 81.4  & 32.1 & 98.3 & \textbf{97.3} & \textbf{98.5} & 87.7& 92.7 & 35.3 & 87.3 & 57.9 & 61.2 & \textbf{100.0} &\textbf{100.0}\\
010\_potted\_meat\_can* &  20.9 & 62.0  & 54.9 & 87.6 & \textbf{82.3} & \textbf{96.6} & \textbf{83.5} & 90.3 & 31.9 & \textbf{92.8} & 59.7 & 77.9 & 64.3 & 81.4\\
011\_banana &  9.9 & 57.7  & 69.1 & 94.0 & \textbf{95.4} & \textbf{98.1} &76.3	& 90.3 & 11.4 & 61.3 & 88.1 & \textbf{100.0} & \textbf{88.3} & 99.3\\
019\_pitcher\_base &  18.1 & 83.7 & 40.4 & 91.1 & \textbf{96.6} & \textbf{97.9} & 86.9 &	92.1  & 6.1 & 88.9 & 93.2 & 99.7 &\textbf{93.9}&\textbf{100.0}\\
021\_bleach\_cleanser & 48.1 & 88.3  & 44.1 & 89.4 & \textbf{93.3} & \textbf{97.4} & 85.5	& 90.8& 32.3 & 89.6 & 85.7 & 91.5 &\textbf{94.3}&\textbf{99.5}\\
024\_bowl & 17.4 & 73.2 & 0.9 & 74.7 & \textbf{89.7} & \textbf{94.9} & 43.6	& 87.5 & 6.7 & 93.2 & 70.4 & 97.9 & 61.5 & \textbf{99.2}\\
025\_mug &  29.5 & 84.8 & 39.2 & 86.5 & \textbf{75.8} & \textbf{96.2} &74.1 &	91.0 & 31.6 & 92.5 & 71.2 & 94.1 &\textbf{92.4}&\textbf{99.8} \\
035\_power\_drill & 12.3 & 60.6  & 19.8 & 73.0 & \textbf{96.3} & \textbf{98.0} & \textbf{96.8}	& 97.0 & 0.0 & 56.0 & 88.4 & 93.9 &93.5&\textbf{99.3}\\
036\_wood\_block &  10.0 & 70.5 & 27.9 & 94.7 & \textbf{94.7} & \textbf{97.4} & 19.9 &	67.1& 0.0 & 77.1 & \textbf{86.0} & \textbf{98.2} & 77.2 & 96.4\\
037\_scissors & 25.0 & 75.5 & 27.7 & 74.2 & \textbf{95.5} & \textbf{97.8} &\textbf{94.7}	& \textbf{97.4} & 0.0 & 75.5 & 78.4 & 89.2 &68.9&85.1\\
040\_large\_marker &38.9 & 81.8 &  74.2 & 97.4 & \textbf{96.5} & \textbf{98.6} & 90.4 &	92.7& 39.3 & 79.6 & \textbf{93.2}& \textbf{97.6} &72.1&84.2\\
051\_large\_clamp & 34.4 & 83.0 & 34.7 & 82.7 & \textbf{92.7} & \textbf{96.9} & 68.9 & 87.4  &  \textbf{100.0} &  \textbf{100.0} & 91.1 & 98.3 &61.2&88.9\\
052\_extra\_large\_clamp &  24.1 & 72.9  & 10.1 & 65.7 & \textbf{94.1} & \textbf{97.6} & 43.7 & 90.5 & 0.0 & 82.6 & \textbf{70.1} & \textbf{90.9} &64.8&89.3\\
061\_foam\_brick &  35.5 & 79.2  & 45.8 & 95.7 & \textbf{93.4} & \textbf{98.1} & 90.9	& 98.7 & 43.5 & 95.2 & 83.8 & 83.9 &\textbf{92.0}&\textbf{98.2}\\
\hline
mean & 24.3 & 71.0 & 42.1 & 88.4 & \textbf{91.5} & \textbf{97.4} & 76.1 &90.4& 25.1 & 86.0 & \textbf{80.6} & 90.3 &78.5&\textbf{95.5}\\
         \bottomrule
    \end{tabular}
    \caption{Full results on the YCB-V dataset. A degradation in ADD AUC is observed for the objects marked with *, although their ADD-S AUC is even higher than that of One2Any. A common characteristic is that these objects are geometrically symmetric but have texture-rich packaging. However, their symmetry definitions are not based on their shape in the evaluation. As a result, the ADD AUC can be low even when the estimated pose is geometrically correct. An  example of geometrically correct pose is presented in \cref{fig:texture_overlap}.} 
    \label{tab:ycbv_data_full}
    \vspace{0.2cm}
\end{table*}

\begin{figure}[!htbp]
    \centering
    \includegraphics[width=0.95\linewidth]{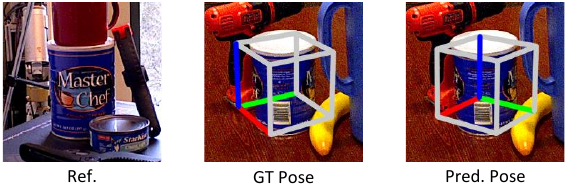}
    \caption{Our model can predict geometrically correct poses under low texture overlap. However, because object symmetry on YCB-V dataset is defined primarily based on texture, such cases can lead to large ADD errors, whereas ADD-S introduces less texture-induced bias in evaluation.}
    \label{fig:texture_overlap}
\end{figure}

\begin{figure}[!htbp]
    \centering
    \includegraphics[width=0.95\linewidth]{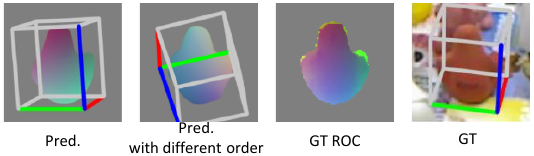}
    \caption{The predicted ROC map may varies under different generation orders. }
    \label{fig:limitation}
\end{figure}

\begin{table*}[]
    \centering
    \footnotesize
    \setlength{\tabcolsep}{0.2mm}
    \begin{tabular}{c|c c|c c|c c|c c||c c| c c | c c}
        \toprule
        \textbf{Method}   & \multicolumn{2}{c|}{\textbf{Wuthrich}} & \multicolumn{2}{c|}{\textbf{RGF}} & \multicolumn{2}{c|}{\textbf{ICG}} & \multicolumn{2}{c||}{\textbf{FoundationPose}} & \multicolumn{2}{c|}{\textbf{FoundationPose}} & \multicolumn{2}{c|}{\textbf{One2Any}} & \multicolumn{2}{c}{\textbf{\ourmodel}}\\
        \hline
         Ref. & \multicolumn{2}{c|}{CAD} & \multicolumn{2}{c|}{CAD}  & \multicolumn{2}{c|}{CAD} & \multicolumn{2}{c||}{16 frames-CAD} & \multicolumn{2}{c|}{$1^{st}$ frame-CAD} & \multicolumn{2}{c|}{$1^{st}$ frame} & \multicolumn{2}{c}{$1^{st}$ frame}\\
        \hline
        metrics of AUC & ADD & ADD-S & ADD & ADD-S & ADD & ADD-S & ADD & ADD-S & ADD & ADD-S & ADD & ADD-S & ADD & ADD-S\\
        \hline
        002\_master\_chef\_can & 55.6 & 90.7 & 46.2 & 90.2 & 66.4 & 89.7 & \textbf{91.2} & \textbf{96.9} & 38.1 & 83.3 & 83.8 & 94.8 & \textbf{92.5} & \textbf{98.7} \\
        003\_cracker\_box & 96.4 & 97.2 & 57.0 & 72.3 & 82.4 & 92.1 & \textbf{96.2} & \textbf{97.5} & 78.3 & 94.0 & 83.0 & 91.3 & \textbf{95.1}& \textbf{98.9} \\
        004\_sugar\_box & 97.1 & 97.9 & 50.4 & 72.7 & \textbf{96.1} & \textbf{98.4} & 94.5 & 97.4 & 40.0 & 78.7 & 88.7 & 95.3 & \textbf{99.3} & \textbf{100.0}\\
        005\_tomato\_soup\_can & 64.7 & 89.5 & 72.4 & 91.6 & 73.2 & 97.3 & \textbf{94.3} & \textbf{97.9} & 14.0 & 49.3 & 87.1 & \textbf{95.5} &\textbf{88.6} &95.4\\
        006\_mustard\_bottle & 97.1 & 98.0 & 87.7 & 98.2 & 96.2 & 98.4 & \textbf{97.3} & \textbf{98.5} & 24.8 & 58.8 & 87.7 & 93.8 & \textbf{92.1}& \textbf{99.8}\\
        007\_tuna\_fish\_can & 69.1 & 93.3 & 28.7 & 52.9 & 73.2 & 95.8 & \textbf{84.0} & \textbf{97.8} & 75.3 & \textbf{97.4} & \textbf{89.5} & 95.9 & 85.8& 97.2\\
        008\_pudding\_box & 96.8 & 97.9 & 12.7 & 18.0 & 73.8 & 88.9 & \textbf{96.9} & \textbf{98.5} & 96.9 & 98.3 & 93.5 & 96.3 &\textbf{100.0}& \textbf{100.0}\\
        009\_gelatin\_box & 97.5 & 98.4 & 49.1 & 70.7 & 97.2 & \textbf{98.8} & \textbf{97.6} & 98.5 & 97.2 & 98.6 & 96.1 & 97.7 &\textbf{100.0}& \textbf{100.0}\\
        010\_potted\_meat\_can & 83.7 & 86.7 & 44.1 & 45.6 & 93.3 & 97.3 & \textbf{94.8} & \textbf{97.5} & 5.5 & 52.6 & \textbf{65.9} & \textbf{84.0} & 62.2 &80.9 \\
        011\_banana & 86.3 & 96.1 & 93.3 & 97.7 & \textbf{95.6} & \textbf{98.4} & \textbf{95.6} & 98.1 & 64.7 & 84.7 & 83.6 & 95.1 &\textbf{91.1}&\textbf{98.5}  \\
        019\_pitcher\_base & 97.3 & 97.7 & 97.9 & 98.2 & \textbf{97.0} & \textbf{98.8} & 96.8 & 98.0 & 94.6 & 96.4 & 87.0 & 93.4 &\textbf{97.8}& \textbf{100.0}\\
        021\_bleach\_cleanser & 95.2 & 97.2 & 95.9 & 97.3 & 92.6 & \textbf{97.5} & \textbf{94.7} & \textbf{97.5} & 16.6 & 58.6 & 84.8 & 93.2 &\textbf{93.1}& \textbf{98.7}\\
        024\_bowl & 30.4 & 97.2 & 24.2 & 82.4 & 74.4 & \textbf{98.4} & \textbf{90.5} & 95.3 & 12.4 & 40.2 & 71.8 & 91.8&\textbf{73.1}&\textbf{98.2}\\
        025\_mug & 83.2 & 93.3 & 60.0 & 71.2 & \textbf{95.6} & \textbf{98.5} & 91.5 & 96.1 & 54.4 & 91.3 & 83.3 & 95.5 &\textbf{96.1}& \textbf{99.8}\\
        035\_power\_drill & 97.1 & 97.8 & 97.9 & 98.3 & 96.7 & 98.5 & \textbf{96.3} & \textbf{97.9} & 50.4 & 69.2 & 85.5 &92.8 &\textbf{96.2}& \textbf{99.6}\\
        036\_wood\_block & 95.5 & 96.9 & 45.7 & 62.5 & \textbf{93.5} & \textbf{97.2} & 92.9 & 97.0 & \textbf{88.4} & \textbf{95.9} & 85.5 & 92.8 &74.9&95.4 \\
        037\_scissors & 4.2 & 16.2 & 20.9 & 38.6 & 93.5 & 97.3 & \textbf{95.5} & \textbf{97.8} & \textbf{96.0} & \textbf{97.9} & 81.0 & 91.7 &69.4& 85.4\\
        040\_large\_marker & 35.6 & 53.0 & 12.2 & 18.9 & 88.5 & 97.8 & \textbf{96.6} & \textbf{98.6} & 74.0 & 90.3 & \textbf{90.3} & \textbf{96.2} &73.0& 83.5\\
        051\_large\_clamp & 61.2 & 72.3 & 62.8 & 80.1 & 91.8 & \textbf{96.9} & \textbf{92.5} & 96.7 & 60.1 & 81.0 & \textbf{84.5} & \textbf{93.2} &75.4& 91.6 \\
        052\_extra\_large\_clamp & 93.7 & 96.6 & 67.5 & 69.7 & 85.9 & 94.3 & \textbf{93.4} & \textbf{97.3} & 44.4 & 85.1 & 71.1 & 91.0 &\textbf{82.5}& \textbf{91.6}  \\
        061\_foam\_brick & 96.8 & 98.1 & 70.0 & 86.5 & 96.2 & \textbf{98.5} & \textbf{96.8} & 98.3 & 89.8 & \textbf{98.2} & \textbf{96.1} & 97.7 &91.5& 97.8\\
        \hline
        mean & 78.0 & 90.2 & 59.2 & 74.3 & 86.4 & 96.5 & \textbf{93.7} & \textbf{97.5} & 57.9 & 80.9 & 84.8 & 93.8 &\textbf{87.1}& \textbf{95.8} \\
        \bottomrule
    \end{tabular}
    \caption{Tracking performance on the YCB-V full video sequences. Results of Wuthrich~\cite{wuthrich2013probabilistic}, RGF~\cite{issac2016depth}, ICG~\cite{stoiber2022iterative}, FoundationPose\cite{wen2024foundationpose}, One2Any\cite{liu2025one2any} are taken from \cite{liu2025one2any}.}
    \label{tab:ycbv_tracking_full}
    \vspace{0.2cm}
\end{table*}

\subsection{Results on Pose Tracking}
We compare the pose tracking performance on the full test sequence of the YCB-V dataset. Following existing methods~\cite{liu2025one2any,wen2024foundationpose}, we use the first frame as the reference for the entire video sequence. As shown in Tab. \ref{tab:ycbv_tracking_full}, our method achieves a higher AUC than existing one-reference methods and shows competitive results against pose tracking approaches based on CAD models. 

        

\subsection{Visualization of the Generation}
\begin{figure*}[!htbp]
    \centering
    \includegraphics[width=0.9\linewidth]{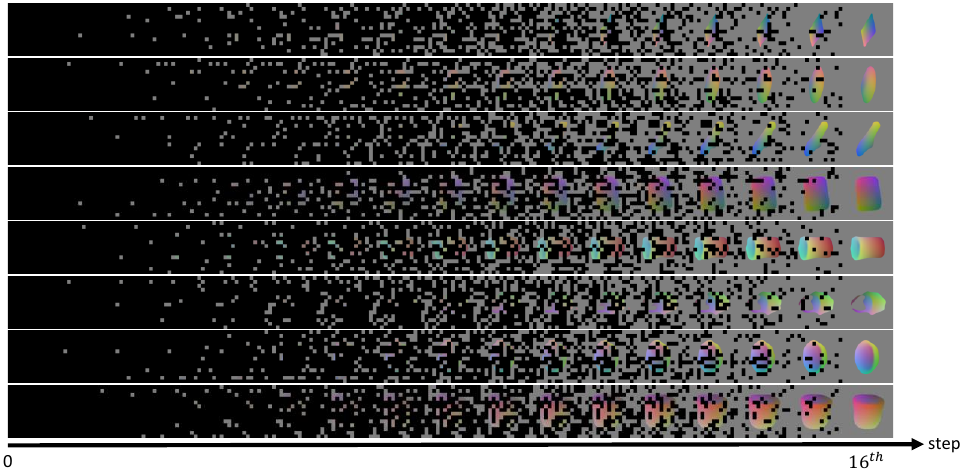}
    \caption{Visualization of the generation. We visualize the generation with 16 steps. Patches corresponding to unpredicted tokens are masked in black.}
    \label{fig:ar_steps}
\end{figure*}

For a better understanding of our method,  we present a visualization of the token generation process. As demonstrated in \cref{fig:ar_steps}, we visualize tokens cumulatively predicted at each step by decoding them using  the tokenizer.

\subsection{Qualitative Results on Toyota-Light}
\begin{figure*}[!htbp]
    \centering
    \includegraphics[width=0.9\linewidth]{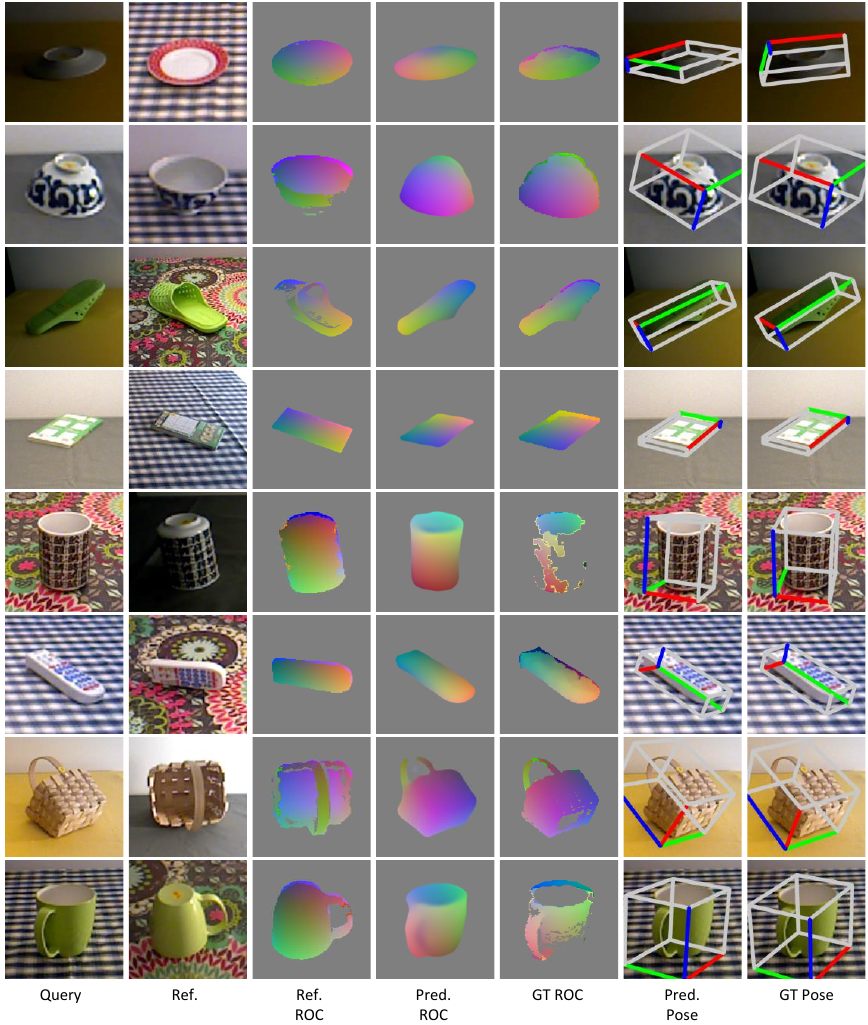}
    \caption{Qualitative results on the Toyota-Light dataset~\cite{hodan2018bop}. }
    \label{fig:tyol}
\end{figure*}

In \cref{fig:tyol}, we present qualitative results on the Toyota-Light dataset~\cite{hodan2018bop}.

\subsection{Qualitative Results on Real-275}
\begin{figure*}[!htbp]
    \centering
    \includegraphics[width=0.9\linewidth]{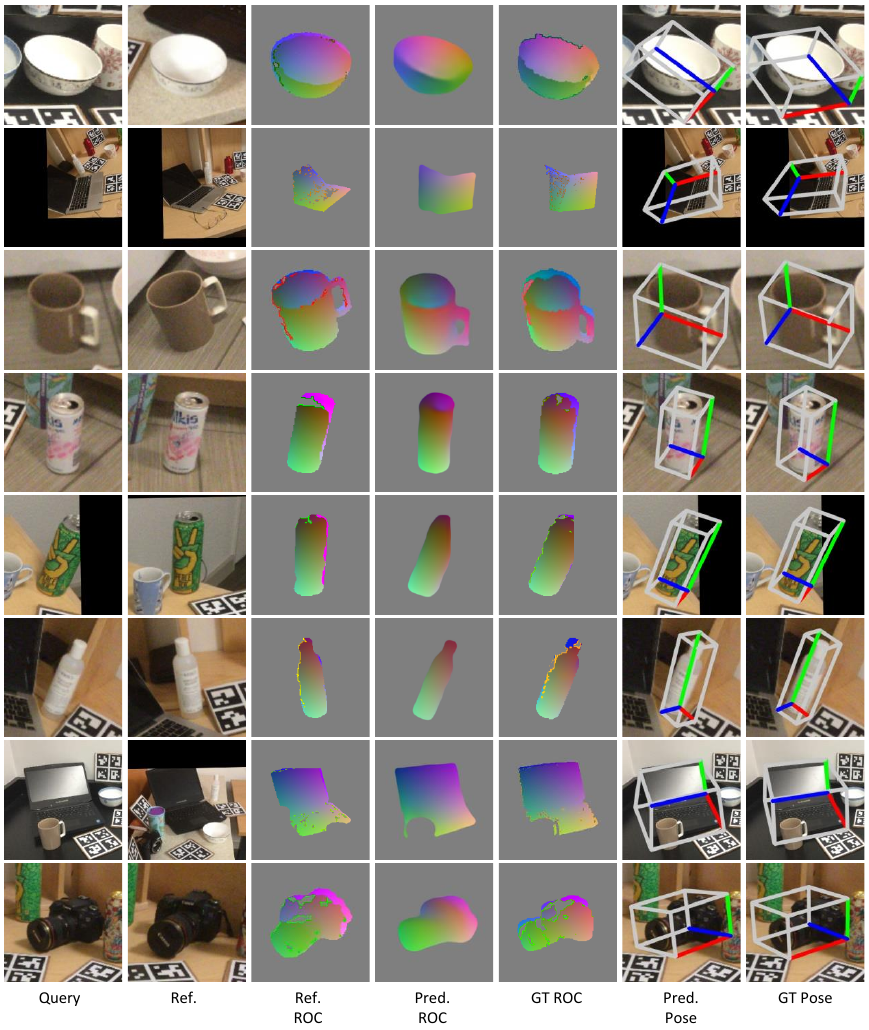}
    \caption{Qualitative results on the Real275 dataset~\cite{wang2019normalized}. }
    \label{fig:real275}
\end{figure*}

In \cref{fig:real275}, we display qualitative results on the Real-275 dataset~\cite{wang2019normalized}.

\subsection{Qualitative Results on  Self-collected Data}
\begin{figure*}[!htbp]
    \centering
    \includegraphics[width=0.9\linewidth]{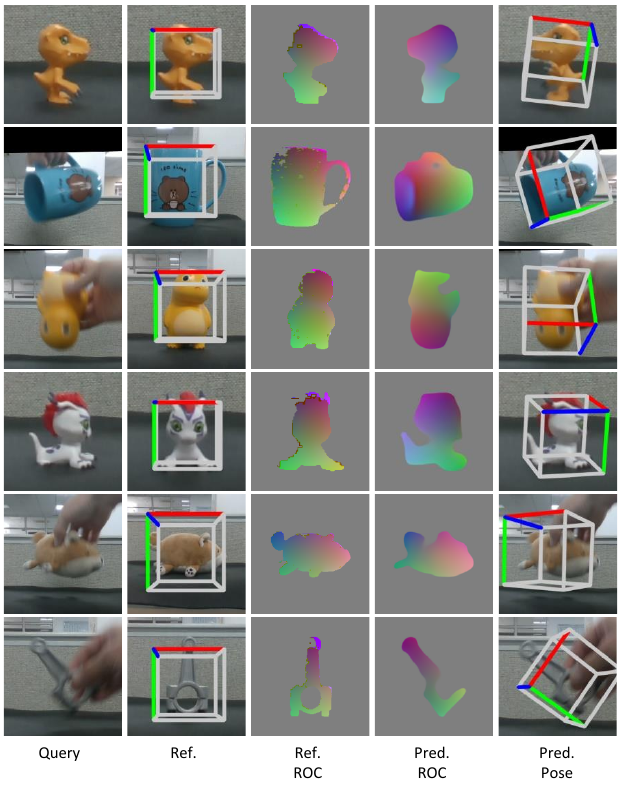}
    \caption{Qualitative results on the self-collected data. Our method demonstrates robust pose estimation capabilities under significant viewpoint variations.}
    \label{fig:self-collected}
\end{figure*}

To further demonstrate the performance of our method on real-world novel objects, we collected RGB-D videos of several common household objects using a RealSense D435i camera. In the first frame, which serves as the reference frame, each object is placed on a platform facing the camera, and its orientation is defined as the canonical rotation. The mask of the object is obtained by Track Anything \cite{yang2023track}.  As shown in \cref{fig:self-collected}, our method accurately estimates the relative poses even when the viewpoints exhibit significant variations.

\section{Limitations}
While demonstrating promising performance, our method is bothered by token generation order. Following previous work~\cite{li2024autoregressive} for image generation, we adopt random order by default and other orders are tried in the supplementary materials. As demonstrated in Fig. \ref{fig:limitation}, we find that generation with an improper token order may result in an erroneous ROC map, particularly when initiating generation from high-uncertainty regions.  This limitation suggests future improvements could be achieved by optimized token ordering strategies, or adoption of next-scale prediction paradigm~\cite{tian2024visual} to mitigate order dependence.

\ifreproStandalone
\FloatBarrier
\bibliography{supp}
\end{document}
\fi

\end{document}